\RequirePackage{amsmath}
\documentclass[runningheads]{llncs}
\usepackage[T1]{fontenc}
\usepackage{graphicx}
\usepackage{subcaption}
\usepackage{xcolor}
\usepackage{pgfplots}
\pgfplotsset{compat=1.8}

\usepackage{adjustbox}
\usepackage{booktabs}
\usepackage{comment}
\usepackage{longtable}

\usepackage{amsmath}
\usepackage{amssymb}
\usepackage{amsfonts}  % For \mathbb
\usepackage{makecell}
\usepackage{url}

\usepackage{listings}

\usepackage{multirow}
\usepackage{hyperref}

\usepackage{cleveref}
\usepackage{caption}
\usepackage{algorithm}
\usepackage{algpseudocode}
\definecolor{lightblue}{RGB}{173,216,230}  % Light Blue
\definecolor{cadetblue}{RGB}{95,158,160}   % Darker Blue~/~Green (Cadet Blue)
\definecolor{lightsalmon}{RGB}{255,160,122} % Light Salmon
\definecolor{indianred}{RGB}{205,92,92}    % Indian Red

\begin{document}
\title{Benchmarking Post-Hoc Unknown-Category Detection in Food Recognition}%\thanks{Supported by organization x.}}
%
%\titlerunning{Abbreviated paper title}
% If the paper title is too long for the running head, you can set
% an abbreviated paper title here
%
\author{Lubnaa Abdur Rahman\inst{1}\orcidID{0000-0001-7271-2814} \and
Ioannis Papathanail\inst{1}\orcidID{0000-0002-7767-3831} \and
Lorenzo Brigato\inst{1}\orcidID{0000-0002-0872-069X} \and
Stavroula Mougiakakou\inst{1}\orcidID{0000-0002-6355-9982}}
\authorrunning{L. Abdur Rahman et al.}
% First names are abbreviated in the running head.
% If there are more than two authors, 'et al.' is used.
%
\institute{ARTORG Center, Graduate School for Cellular and Biomedical Sciences, University of Bern, Bern, Switzerland}
\maketitle              % typeset the header of the contribution
\begin{abstract}
Food recognition models often struggle to distinguish between seen and unseen samples, frequently misclassifying samples from unseen categories by assigning them an in-distribution (ID) label. 
This misclassification presents significant challenges when deploying these models in real-world applications, particularly within automatic dietary assessment systems, where incorrect labels can lead to cascading errors throughout the system. 
Ideally, such models should prompt the user when an unknown sample is encountered, allowing for corrective action. 
Given no prior research exploring food recognition in real-world settings, in this work we conduct an empirical analysis of various post-hoc out-of-distribution (OOD) detection methods for fine-grained food recognition. 
Our findings indicate that virtual logit matching (ViM) performed the best overall, likely due to its combination of logits and feature-space representations.
Additionally, our work reinforces prior notions in the OOD domain, noting that models with higher ID accuracy performed better across the evaluated OOD detection methods.
Furthermore, transformer-based architectures consistently outperformed convolution-based models in detecting OOD samples across various methods.

\keywords{Open Set Food Recognition  \and Out of Distribution \and Image-based Automatic Dietary Assessment }
\end{abstract}
\section{Introduction}

The rise in non-communicable diseases (NCDs), contributing to 74\% of global annual deaths, presents a significant burden on global health \cite{who_ncd}. 
It is well-known that maintaining a healthy diet is key to preventing and managing conditions like malnutrition, cardiovascular diseases, and diabetes while promoting overall well-being \cite{who_emro}. 
To effectively adhere to dietary interventions, continuous and accurate dietary monitoring and assessment are primordial. 
For long, this has been addressed through traditional dietary assessment methods led by dietitians, such as food frequency questionnaires and 24-hour recalls. 
However, not only are they time-consuming but also error-prone due to their reliance on people's memory, introducing subjective biases and inaccuracies \cite{ravelli2020traditional}. 

To overcome these challenges and streamline the process, image-based automatic dietary assessment systems have been proposed by leveraging advancements in artificial intelligence (AI) and computer vision (CV) \cite{vasiloglou2018comparative,lu2020gofoodtm}. 
Such systems have proven effective in multiple settings, such as monitoring malnutrition in hospitalized patients \cite{papathanail2021evaluation}, ensuring adherence to specific diets \cite{papathanail2022feasibility}, and for everyday use under real-life conditions \cite{papathanail2023nutritional}. 
These have further demonstrated accuracy comparable to that of dietitians \cite{lu2020gofoodtm}.
Typically, in an image-based automatic dietary assessment system, a meal image is captured and goes through the different modules of food segmentation, recognition, and volume estimation, all aggregated into the nutritional content of the meal \cite{abdur2023comparative,ABDURRAHMAN202473}. 
One key component of this pipeline is food recognition, whereby the fine-grained label assigned to the food plays a crucial role in the retrieval of nutritional contents from the food composition databases.

%To achieve high accuracy in fine-grained food recognition from images captured in real-world settings, it is essential to train models on datasets that encompass a wide variety of food categories and their diverse representations. However, 
Despite the vast number of potential food classes reflecting the global diversity of foods and eating habits, most food recognition models are practically limited to a predefined set of categories. 
In real-world applications like automatic dietary assessment systems, test samples from user-captured meal images \cite{papathanail2023nutritional} can be out-of-distribution (OOD), meaning that they belong to categories not seen during training.
These OOD samples may present novel food categories or come from entirely different domains, originating from the initial stage of the automatic dietary assessment pipeline, whereby food segmentation models may mistakenly segment non-food items as food \cite{dehais2016food}.

Consequently, when an unknown sample is provided as input, the recognition model is prone to misclassifying this OOD data as an in-distribution (ID) instance.
This misclassification occurs because recognition models are typically trained under the closed-world assumption \cite{bendale2016towards}, where the test data is assumed to be drawn independently and identically distributed (IID) from the same distribution as the training data \cite{yang2024generalized}. 
This problem eventually leads to inaccuracies in the automatic dietary assessment pipeline, which could potentially be dangerous.

For example, in \cite{panagiotou2023complete}, carbohydrate intake coming from an automatic dietary assessment serves as input for the personalized recommendation of insulin dosing for people with diabetes. 
Ideally, the food recognition model should distinguish between what it has not been exposed to and, therefore, prompt an OOD warning to enable the user to assign a proper food label.

Significant advancements have been achieved in the field of open-world multi-class image recognition, particularly in OOD detection \cite{wang2022vim,lee2018simple} and enhancing the applicability of recognition models for open-world scenarios \cite{bendale2016towards}.
However, up to date, there remains a conspicuous gap in modeling these findings to the problem of fine-grained food recognition. 
As such, in this work, our primary contribution is delving into OOD detection for fine-grained food recognition by applying and assessing the performance of several state-of-the-art methods for differentiating between OOD and ID data.
We focus on post-hoc methods, given their advantage of not requiring adjustment to training strategies and simple plug-and-play deployment. 
%These methods are particularly advantageous as they do not require model re-training, thereby saving time and reducing computational costs. 

\section{Related Works} 
\label{sec:related_works}
The core challenge of OOD detection is to create a scoring mechanism that accurately differentiates between ID and OOD samples. 
This is inherently complex due to the unpredictable and vast nature of OOD distributions, which are difficult to model or estimate beforehand. 
There has been increasing interest in developing efficient OOD detection methods that do not require training adjustments, unlike approaches in \cite{lee2017training,zhou2021contrastive}, which can be time- and computationally- consuming. 
Several works focused on establishing post-hoc scoring mechanisms applicable to pre-trained models, which usually assess model confidence using softmax probabilities from the classification layer \cite{hendrycks2016baseline,liang2017enhancing}.
\cite{bendale2016towards} proposed adjusting class probabilities to improve OOD handling as an attempt to make recognition models more ``open'' to real-world scenarios. 
%Similarly, ~\cite{hendrycks2016baseline} introduced the maximum softmax probability (MSP) method to distinguish between in-distribution (ID) and OOD data. 
%Building on MSP, ~\cite{liang2017enhancing} developed the ODIN method, which incorporates temperature scaling and input perturbations to sharpen softmax distributions, making OOD detection more effective.

Other methods leverage feature space metrics by modeling class-conditional distributions within the feature space ~\cite{lee2018simple}. 
%modeled class-conditional distributions in the feature space, using Mahalanobis distance to detect OOD samples by comparing test data to class centroids. 
Logit-based methods have also gained traction as a viable alternative with approaches like ~\cite{hendrycks2019scaling,liu2020energy}.
Other methods proposed combining logit-based while leveraging ~\cite{wang2022vim} representations within the feature space. Some additionally propose modifications to the network's activations during inference, such as clipping or pruning, before applying scoring methods ~\cite{sun2021react,sun2022dice}.
%proposed an approach that scales the logits before applying the softmax function, as well as another method to produce an "anomaly" score. 
%proposed the energy-based model, which computes the negative log-sum-exp of logits to distinguish OOD samples by their lower energy values. 
%To address the issue of OOD detection difficulties in different spaces, ~\cite{wang2022vim} combined a class-agnostic feature space score with ID class-dependent logits.
%In addition to scoring methods, some approaches involve modifications to the network's activations during inference, such as clipping or pruning, aiming to improve OOD detection without retraining the model.
While these methods have been extensively evaluated on standard open-set benchmarks using ID datasets such as CIFAR-10 \cite{krizhevsky2009learning} and ImageNet1K \cite{imagenet_cvpr09}, their applicability to the fine-grained food recognition, remains unexplored.
%Fine-grained recognition often present unique challenges \cite{krause2014learning}, including higher intra-class variability and the presence of visually similar classes, which may further complicate OOD detection. 

\section{Materials and Methods}
Our task here is to first train different models for multi-class fine-grained food recognition and apply the existing different post-hoc OOD methods to further assess their performance. 
All the experiments described in this section, including \Cref{sec:llama3,sec:training,sec:oodmethods}, were run with acceleration on NVIDIA RTX A6000 with the exception of the training of the Shifted
Window - T variant (Swin-T) \cite{liu2021Swin} which was on RTX 4090\footnote{Computations were performed on UBELIX (https://www.id.unibe.ch/hpc), the HPC cluster at the University of Bern.}. 

\subsection{Datasets}
\subsubsection{In distribution}

For the ID dataset to train the model on, we utilize the Food-101 dataset \cite{bossard2014food}, a well-known fine-grained food recognition dataset comprising 101,000 images categorized into 101 distinct food classes covering diverse cuisines. 
We use the standard train/test splits of 75,750/25,250 images. 

\subsubsection{Out of distribution}
\begin{figure}
\centering
\includegraphics[width=0.9\textwidth]{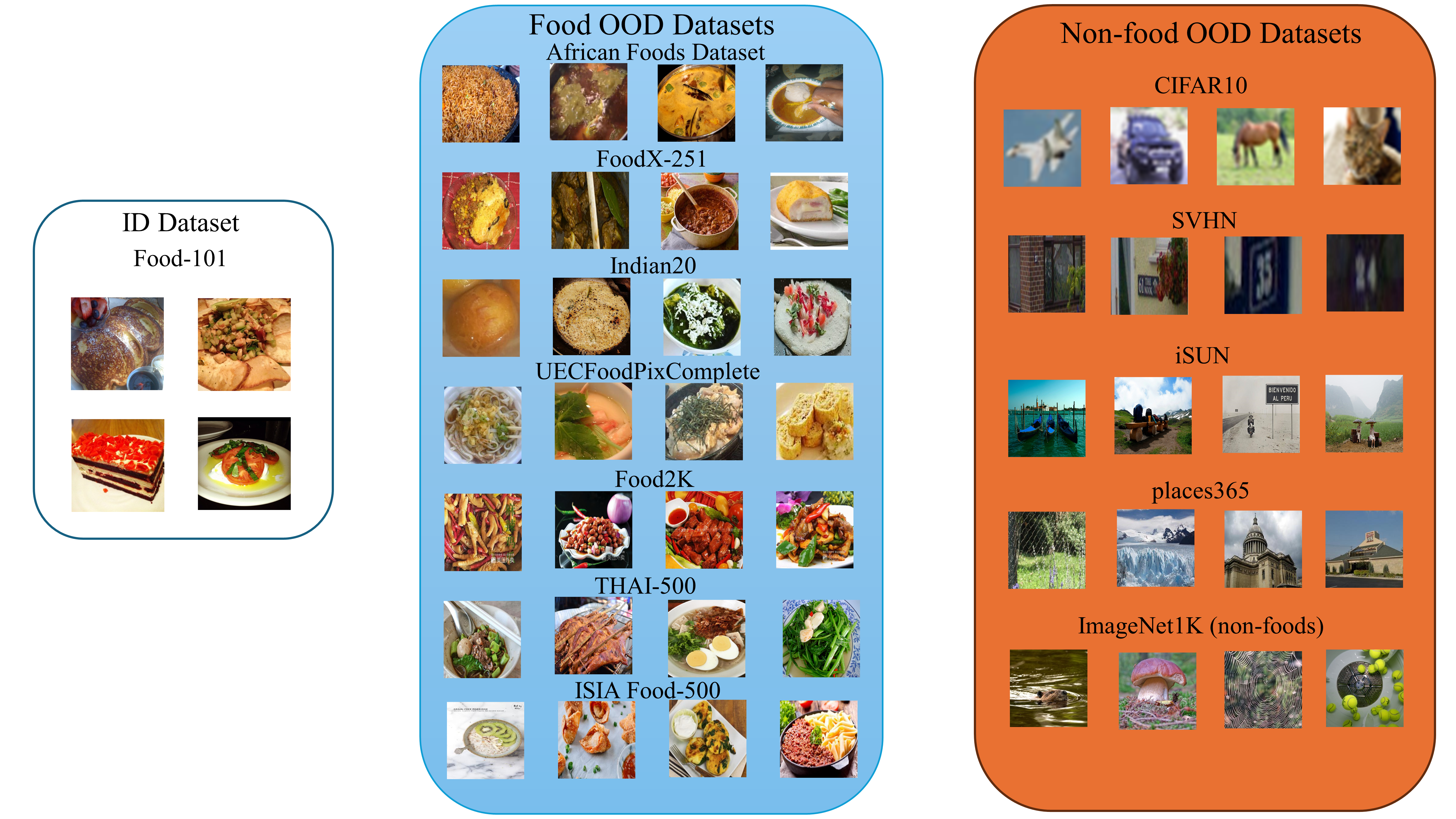}
\caption{Example of images present in ID and OOD datasets} \label{fig1}
\end{figure}

%When it comes to OOD detection for food recognition, the model should be capable of identifying as OOD not only food samples that do not belong to the distribution on which it was trained but also non-food items. 
\paragraph{Food datasets}
As no prior work exists on OOD detection within the food domain and no benchmark OOD datasets are available for this task, we, therefore, utilize the open-source food recognition datasets presented in \Cref{tab:dataset_overview}.
Overlapping and similar categories are removed (as described in \Cref{sec:llama3}).
We also provide the list of categories used for each dataset in the Supplementary Materials: Supplementary A - OOD. 
Specifically for Indian20 \cite{pandey2022object} and UECFoodPixComplete \cite{uecfoodpixcomplete}, since there were multiple foods per image, we further used annotations provided by the dataset creators to produce single food item images.
We show some examples of the images from each dataset in \Cref{fig1}.

\begin{table}[h!]
\renewcommand{\arraystretch}{0.8}
\caption{Food Dataset Overview and Image Counts}
\label{tab:dataset_overview}
\centering
\begin{tabular}{lccc} 
\toprule
\makecell{\textbf{Dataset}} & \makecell{\textbf{Cuisine}} & \makecell{\textbf{\#Images}} \\
\midrule
\textbf{African Foods Dataset} \cite{ataguba2024african} & African (Cameroon, Ghana)     & 1,751 \\

\textbf{FoodX-251} \cite{kaur2019foodx}     & Mixed Cuisines                & 6,561 \\
\textbf{Indian20} \cite{pandey2022object}   & Indian                       & 37,067 \\
\textbf{UECFoodPixComplete} \cite{uecfoodpixcomplete}   & Japanese                    & 8,758 \\
\textbf{Food2K} \cite{min2023large}         & Mixed Cuisines                & 262,777 \\
\textbf{THFOOD-50} \cite{termritthikun2018nu} & Thai                         & 1,534 \\
\textbf{ISIA FOOD-500} \cite{min2020isia}   & Mixed Cuisines                & 98,482 \\
\bottomrule
\end{tabular}
\end{table}

\paragraph{Non-Food datasets}
To compare with common benchmarks and address scenarios specifically in dietary assessment pipelines where segmentation models may inaccurately segment non-food items as well, we use the OOD datasets presented in \Cref{tab:non_food_dataset_overview}. 
For all datasets, we utilized the testing splits, except for the ImageNet1K dataset \cite{imagenet_cvpr09}, where we used the validation set after filtering it to include only non-food items.
We use the ImageNet1K to also evaluate whether pretraining affects OOD detection in the ResNet-18, ResNet-50 \cite{he2016deep}, and Swin-T \cite{liu2021Swin} models.
Examples of images from each dataset are shown in \Cref{fig1}.

\begin{table}[h!]
\renewcommand{\arraystretch}{0.8}
\caption{Non-Food Dataset Overview and Image Counts}
\label{tab:non_food_dataset_overview}
\centering
\begin{tabular}{lcc} 
\toprule
\makecell{\textbf{Dataset}} & \makecell{\textbf{Type}} & \makecell{\textbf{\#Images}} \\
\midrule
\textbf{CIFAR10} \cite{krizhevsky2009learning}   & Animals \& Vehicles & 10,000 \\
\textbf{iSUN} \cite{xu2015turkergaze}            & Scenes        & 8,925 \\
\textbf{ImageNet1K} (non-food items) \cite{imagenet_cvpr09} & Diverse & 47,150 \\
\textbf{Places365} \cite{zhou2017places}         & Scenes       & 328,500 \\
\textbf{SVHN} \cite{netzer2011reading}           & Street View House Numbers  & 26,032 \\
\bottomrule
\end{tabular}
\end{table}

\begin{comment}
\begin{itemize}
\item \textbf{CIFAR10} \cite{krizhevsky2009learning} with 10,000 images.
\item \textbf{SVHN} \cite{netzer2011reading} with 26,032 images.
\item \textbf{Places365} \cite{zhou2017places} with 328,500 images.
\item \textbf{iSUN} \cite{xu2015turkergaze}: This dataset is a subset of the SUN dataset consisting of 8,925 images.
\item \textbf{ImageNet1K} (non-food items) \cite{imagenet_cvpr09}: We used non food items from the validation set, which included 47,150 images. We decided to use this dataset also to understand to what extent the pre-training, specifically in the resnet50, and resnet18 models, affected the OOD detection.

\end{itemize}
\end{comment}
 %LSUN-Crop [66],
%LSUN-Resize [66], and iSUN [64]

\subsubsection{Removal of Overlapping and Similar Categories}
\label{sec:llama3}
To remove overlapping categories, we compared word-for-word patterns between the OOD datasets and the ID dataset, eliminating any matching entries from the OOD datasets. 
Additionally, given the complexity of lexicons, including synonyms and words with different meanings across languages, we further processed the categories of the OOD datasets. 
For this task, we used Llama-3-8B-Instruct \cite{llama3modelcard}. 
We instructed Llama-3 to act as a global cuisine expert to identify whether any OOD categories were synonymous or represented similar categories to those in the in-distribution dataset. 
For example, ``Beef tenderloin'' from Food2k was removed because it encompasses the ID category ``Filet mignon'' from Food-101.

\subsection{Training}
\label{sec:training}
We evaluated four models, chosen based on recent works \cite{wang2022vim,oh2024we}, two CNN-based, ResNet18 
and ResNet50 \cite{he2016deep}, and two transformer-based architectures, the Vision Transformer with B variant patch 16  (ViT-B/16) \cite{dosovitskiy2020image}, and the Swin-T\cite{liu2021Swin}. 
All models were trained with a batch size of 64, and the images were resized to 224$\times$224.
The training specifics and ID accuracy achieved by the different models are presented in \Cref{tab:Training specifics}. 
To compare OOD detection performance between fine-tuned models and models trained from scratch (-S), we focused on ResNet-18 and ResNet-50.

\begin{table}[h!]
\renewcommand{\arraystretch}{0.8}
\caption{Training Specifics and ID Performance}
\label{tab:Training specifics}
\centering
\resizebox{1.0\textwidth}{!}{ 
\begin{tabular}{lcccccccc} % Adjusted to 9 columns
\toprule
\makecell{\textbf{Model}} & \makecell{\textbf{Pre-training}} & \makecell{\textbf{Scheduler}} & \makecell{\textbf{Optimizer}} & \makecell{\textbf{LR}} & \makecell{\textbf{WD}} & \makecell{\textbf{Epochs}} & \makecell{\textbf{Accuracy}} \\
\midrule
ResNet-18  & ImageNet-1K  & Linear & SGD & 0.001   & 0  & 50  & 77.71\% \\
ResNet-18-S  & None     & Cosine & SGD & 0.1     & 0.0001  & 100 & 65.04\% \\
ResNet-50   & ImageNet-1K  & Cosine & SGD & 0.001   & 0.0001  & 50  & 82.91\% \\
ResNet-50-S & None     & Cosine & SGD & 0.1     & 0.0001  & 100 & 67.62\% \\
ViT-B      & ImageNet-21K  & Cosine & SGD & 0.001   & 0       & 15  & \underline{\textbf{89.34\%}} \\
Swin-T      & ImageNet-1K  & Linear & Adam & 0.00001   & 0       & 15  & 87.73\% \\
\bottomrule
\end{tabular}
}
\end{table}

\subsection{Post-hoc OOD Detection}
\subsubsection{Preliminaries}
We begin by outlining the general framework whereby the input space can be denoted by \( \mathcal{X} = \mathbb{R}^d \) and the output space by \( \mathcal{Y} = \{1, 2, \dots, C\} \). 
The model is provided with a training set \( \mathcal{D} = \{(\mathbf{x}_i, y_i)\}_{i=1}^N \), sampled from an unknown joint distribution \( P(\mathcal{X}, \mathcal{Y}) \).
Let \( P_{\text{in}} \) represent the marginal probability distribution on \( \mathcal{X} \). 
We recall that the principle of OOD detection is traditionally modeled as a binary classification problem whereby the ID is considered as the positive class and the OOD the negative class. 
At a specific evaluation time, the goal is to determine whether an input sample \( \mathbf{x} \in \mathcal{X} \) belongs to the ID \( P_{\text{in}} \) or the OOD \( P_{\text{out}} \). 
The distribution of \( P_{\text{out}} \) represents unknowns that may be encountered during deployment, whereby the label set has no overlap with ID labels \( \mathcal{Y} \) and should not be predicted by the model. 
As such, post-hoc OOD detection follows the pipeline as expressed in \Cref{alg:recogition_model_pseudo}.
\( \lambda \) represents the cutoff threshold to distinguish between ID and OOD samples.
Following common practice, we choose \(\lambda\) such that 95\% of the ID data is correctly classified.

\begin{algorithm}
\caption{Recognition Model with Post-hoc OOD Detection}
\label{alg:recogition_model_pseudo}
\begin{algorithmic} 
\Function{PostHocOODDetection}{trained\_model, deployment\_data}
   \For{each sample $\mathbf{x}$ in deployment\_data}
       \State Extract features $\mathbf{f}(\mathbf{x})$ from the input using the model's feature extractor
       \State Compute the score $S(\mathbf{f}(\mathbf{x}))$ using a scoring function
       % \State $S(\mathbf{f}(\mathbf{x})) \gets \text{ScoringFunction}(trained\_model, \mathbf{f}(\mathbf{x}))$
        
      \If{$S(\mathbf{f}(\mathbf{x})) \geq \lambda$}
          \State label $\gets$ \text{``in''}  \Comment{ID sample}
          \State predicted\_label $\gets$ \text{trained\_model.classifier}($\mathbf{f}(\mathbf{x})$) 
       \Else
           \State label $\gets$ \text{``out''}  \Comment{OOD sample}
           \State predicted\_label $\gets$ \text{``unknown''}
       \EndIf
        
      \State \textbf{output} predicted\_label
  \EndFor
\EndFunction
\end{algorithmic}
\end{algorithm}

\subsubsection{Benchmarked OOD Methods}
\label{sec:oodmethods}
In this section, we describe the post-hoc OOD baselines evaluated in this work.
For the entailing section, let \( z = [z_1, z_2, \dots, z_C] \in \mathbb{R}^C \) be the logits for a sample, where \( C \) is the number of ID classes and \( z_i \) represents the logit for class \( i \).

\paragraph{Maximum Softmax Probility \cite{hendrycks2016baseline}:} The maximum softmax probability score (MSP) is based on the observation that correctly classified ID examples tend to have higher softmax probabilities than OOD examples. The softmax probability for class \( i \) is given by:
\(
p(y = i | x) = \frac{\exp(z_i)}{\sum_{j=1}^{C} \exp(z_j)}
\). The MSP score \( MSP(x) \) is then computed as:
\begin{equation}
MSP(x) = \max_{i} p(y = i | x)
\label{eq:msp}
\end{equation}

\paragraph{ODIN \cite{liang2017enhancing}:} The Out-of-DIstribution detector for Neural networks (ODIN) method applies temperature scaling to adjust the softmax scores, enhancing the separation between ID and OOD data.
The softmax score \( S_i(x; T) \) for class \( i \) at temperature \( T \) is given by:
\begin{equation}
S_i(x; T) = \frac{\exp(z_i / T)}{\sum_{j=1}^{C} \exp(z_j / T)}
\label{eq:odin}
\end{equation}
In our experiments, we set \( T \) to 1000 as in previous work \cite{liang2017enhancing,hsu2020generalized}.

\paragraph{OpenMax \cite{bendale2016towards}:} OpenMax replaces the standard softmax layer with a layer that recalibrates the output probabilities by modeling the uncertainty for each class by fitting a Weibull distribution,
\( W(\lambda_i, k_i) \) for each class \( i \). 
Here, \( \lambda_i \) and \( k_i \) are the scale and shape parameters of the Weibull distribution, respectively for each class. 
The original logits \( z_i \) are adjusted by reducing the score for the top classes based on the tail probability that they belong to an unknown class computed by \(\
z'_i = z_i \cdot \left(1 - W(\lambda_i, k_i, z_i)\right)
\), where \( W(\lambda_i, k_i, z_i) \) represents the Weibull cumulative distribution function for class \( i \), evaluated at logit \( z_i \). 
After adjusting the logits, a recalibrated softmax probability distribution is computed using the adjusted logits \( z' \). The probability that the input belongs to a known class is then given by:
\begin{equation}
p(y = i | x) = \frac{\exp(z'_i)}{\sum_{j=1}^{C} \exp(z'_j) + \exp(z'_\alpha)}
\label{eq:openmax}
\end{equation}
where \( z'_\alpha \) is the score for the unknown class, which is derived from the tail probabilities of the Weibull distributions.

\paragraph{KL Matching \cite{hendrycks2019scaling}:} 
Kullback-Leibler (KL) Matching captures the typical shape of each class's posterior distribution by creating class-wise posterior distribution templates. It compares the network's softmax posterior distribution \( p(y | x) \) to the class-wise template distributions \( q(y | c) \), where \( c \) is the class template. The anomaly score \( KL(x) \) is computed as:\begin{equation}
KL(x) = \min_{c} D_{\text{KL}}\left( p(y | x) \parallel q(y | c) \right)
\label{eq:kl_matching}
\end{equation}
where \( D_{\text{KL}} \) is the Kullback-Leibler divergence. A higher \( KL(x) \) score indicates a greater likelihood that the input is OOD, as the posterior distribution of the input deviates significantly from the class-wise templates.

\paragraph{Mahalanobis \cite{lee2018simple}:}
This method leverages Gaussian distributions %\cite{goodman1963statistical} 
to model the class-conditional distributions resulting in a Mahalanobis distance-based scores. 
By estimating the generative classifier parameters from a pre-trained softmax neural classifier, the empirical class means and covariances are calculated using the activations on training data.
Let \( \mu_i \) represent the empirical mean vector for class \( i \), and let \( \Sigma \) represent the shared covariance matrix, both of which were estimated using the training set. 
%These parameters are estimated from the output of the penultimate layer of the trained model. 
Given a new input test sample \( x \), the Mahalanobis distance between the feature representation \( f(x) \) and the class mean \( \mu_i \) is given as:
\(
d_M(f(x), \mu_i) = (f(x) - \mu_i)^T \Sigma^{-1} (f(x) - \mu_i).
\)
This quantity measures how far the feature representation of the input \( f(x) \) is from the mean of the distribution for class \( i \). 
To detect OOD data, the confidence score, \(M(x)\), is:

\begin{equation}
M(x) = \max_{i} \left\{- d_M(f(x), \mu_i)\right\}
\label{eq:Mahalanobis}
\end{equation}
Inputs exhibiting larger \(M(x)\) scores (i.e., lower confidence scores) are more likely to be classified as OOD.

\paragraph{Maximum Logit \cite{hendrycks2019scaling}:} The MaxLog score uses the negative of the maximum of the unnormalized logits as an anomaly score which is given by:
\begin{equation}
MaxLog(x) = - \max_{i} z_i
\label{eq:maxlog}
\end{equation}
%After computing the MaxLog score for each input test sample, the thresholding can be applied as shown in \Cref{alg:recogition_model_pseudo}.

\paragraph{Energy \cite{liu2020energy}:} 
The Energy score maps each input to a single non-probabilistic scalar that is
lower for ID data and higher for OOD ones. It is computed as the negative log-sum-exp of the logits, effectively distinguishing OOD samples based on their lower energy values compared to ID data. 
The Energy score \( E(x) \) for a given input \( x \) is:
\begin{equation}
E(x) = - \log \left( \sum_{i=1}^{C} \exp(z_i) \right)
\label{eq:energy}
\end{equation}
%After computing the Energy score for each input test sample, the thresholding can be applied as shown in \Cref{alg:recogition_model_pseudo}.

\paragraph{ViM \cite{wang2022vim}:} 
Virtual-logit Matching (ViM) combines a class-agnostic score from the feature space with the ID class-specific logits. 
First, the feature space offset using vector \( o = -\left(W^T\right)^+ b\), where \( W \) and \( b \) are the weights and biases from the fully connected layer. 
The principal subspace \( P \) is obtained via eigen decomposition on the training data (in our experiments, we use the whole training set). 
The residual \( x_{P^\perp} \) is then computed as the projection of feature \( x \) onto the orthogonal complement of \( P \). 
Next, the norm of the residual \( x_{P^\perp} \) is scaled to create a virtual logit, \( l_0 := \alpha \| x_{P^\perp} \| = \alpha \sqrt{x^\top R R^\top x} \), where \( \alpha \) is a scaling factor derived from the training data and \( R \) represents the matrix whose columns span the orthogonal complement of the principal subspace \( P \). 
The virtual logit \( l_0 \) is then appended to the original logits \( z \), forming a new logit vector \( [z_1, z_2, \dots, z_C, l_0] \).
%, and the softmax function is applied to the combined logits. 
The ViM score is then given as the softmax probability corresponding to the virtual logit:
\begin{equation}
ViM(x) = \frac{e^{l_0}}{\sum_{i=1}^C e^{z_i} + e^{l_0}}
\label{eq:ViM}
\end{equation}

\paragraph{ReAct \cite{sun2021react}:} 
Rectified Activation (ReAct), designed to reduce the model's overconfidence, works by applying rectifications to the activations applied to the penultimate layer of the network and truncates activations to a certain limit, \(\tau\).
%, to minimize noise whereby higher value of \(\tau\) represents a larger threshold of activation truncation. 
After truncation, the outputs can be combined with the \Cref{eq:energy} scoring function for OOD detection. 
In our experiments, we run ReAct with different values of \(\tau\) and show the best results in \Cref{tab:OODFOOD,tab:OODNF}.

\paragraph{DICE \cite{sun2022dice}:} 
Directed Sparsification (DICE) introduces sparsity into the activations of the penultimate layer using a contribution matrix derived from the whole training set to retain the most significant weights. 
Sparsity is applied by zeroing out a certain percentage of the lowest activations based on a sparsity parameter 
\(\rho\) where higher values of \(\rho\) correspond to a larger proportion of weights being dropped. 
Once the ``unimportant'' weights are removed, scoring functions such as \Cref{eq:energy} or \Cref{eq:msp} can be utilized. 
In our experiments, consistent with \cite{sun2022dice}, \Cref{eq:energy} provided the best results. 
Various values of \(\rho\) were tested, and the optimal value for each model is reported in \Cref{tab:OODFOOD,tab:OODNF}.

\subsubsection{Evaluation Metrics}
We used two key metrics for evaluating OOD detection methods:

\begin{itemize}
\item {FPR95:} The false positive rate (FPR) at 95\% true positive rate (TPR), which is computed as:
\(
\text{FPR95} = P(\text{False Positive} \mid \text{TPR} = 95\%)
\).

\item{Area Under the Receiver Operating Characteristic Curve (AUROC):} It is defined as:
\(
\text{AUROC} = \int_0^1 \text{TPR} \, d\text{FPR}
\).
\end{itemize}

\section{Results and Discussion}
\begin{figure}[ht]
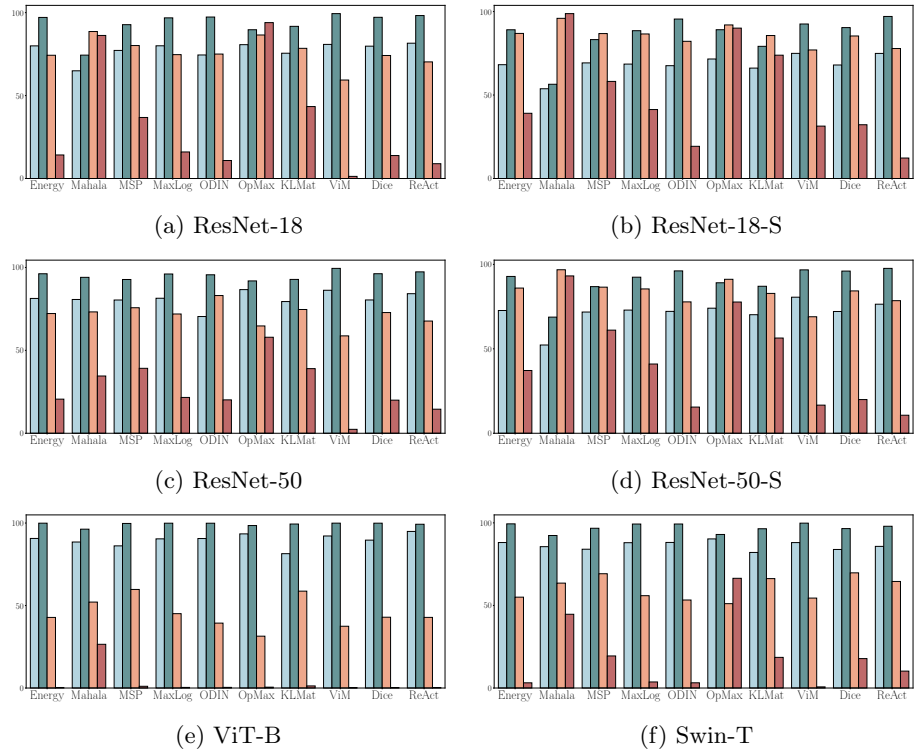

    \centering
    
    % First row: ResNet-18 and ResNet-18-S
    \begin{subfigure}{0.49\textwidth}
        \resizebox{\linewidth}{!}{\input{Resnet18_pg_plot.pgf}}
        \caption{ResNet-18}
        \label{fig:resnet18}
    \end{subfigure}
    \hfill
    \begin{subfigure}{0.49\textwidth}
        \resizebox{\linewidth}{!}{\input{Resnet18Scratch_pg_plot.pgf}}
        \caption{ResNet-18-S}
        \label{fig:resnet18s}
    \end{subfigure}
    
    % Second row: ResNet-50 and ResNet-50-S
    \vspace{0.1cm} % Space between rows
    \begin{subfigure}{0.49\textwidth}
        \resizebox{\linewidth}{!}{\input{Resnet50_pg_plot.pgf}}
        \caption{ResNet-50}
        \label{fig:resnet50}
    \end{subfigure}
    \hfill
    \begin{subfigure}{0.49\textwidth}
        \resizebox{\linewidth}{!}{\input{Resnet50Scratch_pg_plot.pgf}}
        \caption{ResNet-50-S}
        \label{fig:resnet50s}
    \end{subfigure}
    
    % Third row: ViT-B and Swin-T
    \vspace{0.1cm} % Space between rows
    \begin{subfigure}{0.49\textwidth}
        \resizebox{\linewidth}{!}{\input{ViT-B_pg_plot.pgf}}
        \caption{ViT-B}
        \label{fig:vitb}
    \end{subfigure}
    \hfill
    \begin{subfigure}{0.49\textwidth}
        \resizebox{\linewidth}{!}{\input{SWIN_T_pg_plot.pgf}}
        \caption{Swin-T}
        \label{fig:swint}
    \end{subfigure}
    
    % Caption for the entire figure
    \caption{Average results for different methods for different models for food and non-foods. The color-coded metrics are as follows: 
    \colorbox{lightblue}{AUROC for foods}, 
    \colorbox{cadetblue}{AUROC for non-foods},
    \colorbox{lightsalmon}{FPR95 Foods}, 
    and \colorbox{indianred}{FPR95 for non-foods}.}
    \label{fig:all_models}
\end{figure}

\begin{table}[ht]
\caption{Detailed performance for Food OOD dataset expressed by FPR95~/~AUROC. Best results are represented by \underline{\textbf{FPR95}}.}
\label{tab:OODFOOD}
\centering
\resizebox{\textwidth}{!}{ % Resize table to fit text width
%%OLD
\begin{tabular}{c|c|c|c|c|c|c|c|c|c}
\toprule
%Model & Method & African & FoodX-251 & Indian20 & UECFoodPixC & Food2K & THFOOD-50 & ISIA Food-500 & Average \\
\textbf{Model} & \textbf{Method} & \textbf{African} & \textbf{FX251} & \textbf{Indian} & \textbf{UECFPC} & \textbf{Food2K} & \textbf{THF50} & \textbf{ISIA F500} & \textbf{Average} \\
\midrule
\multirow{10}{*}{RN18} 
                       & MSP                   & 77.1~/~79.3   & 79.7~/~78.2  & 79.9~/~77.7   & 79.7~/~76.8  & 79.9~/~77.7  & 81.5~/~76.9  & 83.3~/~73.9   & 80.2~/~77.2    \\
                       %\cline{2-10}
                       & ODIN                  & 76.6~/~73.7  & 76.1~/~76.2  & 63.7~/~77.9  & 75.8~/~71.2  & 71.5~/~78.3  & 82.7~/~71.4  & 78.7~/~72.6   & 75.0~/~74.5    \\ %\cline{2-10}
                       & OpenMax               & 81.7~/~83.4   & 91.7~/~78.5  & 83.7~/~82.9  & 85.0~/~81.6  & 87.9~/~81.2  & 87.2~/~80.2  & 88.5~/~77.0   & 86.5~/~80.7    \\ %\cline{2-10}
                       & KL Matching           & 74.2~/~78.2  & 79.9~/~76.5  & 77.7~/~76.9  & 77.6~/~75.1  & 77.4~/~75.5   & 80.4~/~74.4  & 82.1~/~72.1    & 78.5~/~75.5    \\ %\cline{2-10}
                       & Mahalanobis           & 82.5~/~68.8  & 92.3~/~61.9   & 79.6~/~73.3  & 89.1~/~67.0   & 91.9~/~62.2   & 93.8~/~59.2  & 91.1~/~62.1   & 88.6~/~64.9    \\ %\cline{2-10}
                       & MaxLog                & 69.3~/~84.0  & 75.3~/~80.6   & 73.8~/~80.8    & 73.4~/~79.2  & 73.3~/~81.4   & 77.7~/~78.4  & 79.9~/~75.7   & 74.7~/~80.0    \\ %\cline{2-10}
                       & Energy                & 69.1~/~84.1  & 75.0~/~80.5  & 73.1~/~80.8  & 72.9~/~79.1   & 72.5~/~81.4  & 78.0~/~78.2  & 79.7~/~75.5   & 74.3~/~80.0    \\ %\cline{2-10}
                       & ViM                   & 55.6~/~81.4  & 64.2~/~82.5  & 53.6~/~84.6  & 61.3~/~78.9   & 54.7~/~82.5  & 58.5~/~79.8  & 67.7~/~76.3   & \underline{\textbf{59.4}}~/~80.8    \\ %\cline{2-10}
                       & ReAct \(\tau\)=1.75   & 59.0~/~87.1  & 74.3~/~80.7  & 64.3~/~84.0  & 68.3~/~81.5   & 69.9~/~82.5  & 78.2~/~78.8   & 78.2~/~76.4   & 70.3~/~81.6    \\ %\cline{2-10}
                       & DICE \(\rho\)=0.01       & 68.4~/~83.9  & 75.2~/~80.2   & 73.2~/~80.4  & 72.8~/~79.0  & 72.4~/~81.3  & 77.2~/~78.4  & 80.0~/~75.3   & 74.2~/~79.8    \\ \cmidrule{1-10}

\multirow{10}{*}{RN50} 
                       & MSP                   & 73.5~/~82.5   & 98.5~/~77.0  & 80.9~/~94.2   & 73.2~/~81.9  & 79.1~/~75.1  & 79.7~/~91.6   & 75.4~/~81.2   & 75.7~/~80.4    \\ %\cline{2-10}
                       & ODIN                  & 86.0~/~68.9  & 96.7~/~86.1  & 70.2~/~89.6  & 74.8~/~73.3   & 64.6~/~84.1  & 66.1~/~84.1    & 78.8~/~76.0   & 83.1~/~70.4    \\ %\cline{2-10}
                       & OpenMax               & 50.3~/~90.5  & 99.2~/~74.3   & 84.3~/~95.5  & 66.2~/~87.8   & 87.1~/~62.8  & 87.3~/~95.1  & 64.1~/~87.1   & 64.7~/~86.7    \\ %\cline{2-10}
                       & KL Matching           & 72.1~/~81.8  & 98.3~/~75.8  & 80.2~/~93.4  & 72.1~/~81.1  & 74.2~/~74.4    & 79.1~/~90.8  & 74.1~/~80.1   & 74.7~/~79.4    \\ %\cline{2-10}
                       & Mahalanobis           & 56.7~/~87.8  & 99.0~/~78.5   & 79.3~/~93.8  & 67.2~/~83.8  & 80.7~/~73.2  & 81.2~/~92.3   & 75.7~/~79.9   & 73.2~/~80.7    \\ %\cline{2-10}
                       & MaxLog                & 69.1~/~84.1  & 98.6~/~73.2   & 81.6~/~94.1  & 66.8~/~83.6  & 79.1~/~70.8  & 80.9~/~91.7  & 70.7~/~83.0   & 71.9~/~81.4    \\ %\cline{2-10}
                       & Energy                & 69.7~/~84.1  & 98.6~/~73.4  & 81.4~/~94.1  & 66.6~/~83.5  & 79.1~/~70.8  & 80.8~/~91.6  & 70.9~/~83.0   & 77.4~/~81.3    \\ %\cline{2-10}
                       & ViM                   & 36.5~/~92.7  & 99.4~/~66.4  & 85.3~/~95.6  & 37.8~/~91.8   & 89.2~/~57.1  & 86.5~/~94.2  & 64.2~/~85.9   & \underline{\textbf{58.7}}~/~86.3    \\ %\cline{2-10}
                       & ReAct \(\tau\)=1      & 57.7~/~88.1  & 99.1~/~72.4   & 84.1~/~94.4  & 61.5~/~85.1  & 84.2~/~63.4  & 85.5~/~94.7   & 67.8~/~87.9   & 70.0~/~85.1    \\ %\cline{2-10}
                       & DICE \(\rho\)=0.05       & 68.3~/~84.0  & 98.6~/~73.3   & 81.8~/~94.1  & 66.7~/~83.6  & 79.3~/~70.8  & 80.9~/~91.6   & 70.7~/~83.0   & 72.3~/~81.5    \\ \cmidrule{1-10}

\multirow{10}{*}{ViT-B} 
                       & MSP                   & 56.8~/~86.6  & 62.5~/~86.1  & 55.5~/~88.0  & 62.8~/~84.4  & 55.8~/~88.0  & 61.1~/~85.2  & 63.8~/~84.9  & 59.8~/~86.2    \\ %\cline{2-10} 
                       & ODIN                  & 33.6~/~92.3  & 45.1~/~89.8  & 30.0~/~93.4  & 41.7~/~89.1  & 34.3~/~92.3  & 44.0~/~88.7  & 46.7~/~88.7  & 39.3~/~90.6    \\ %\cline{2-10} 
                       & OpenMax               & 26.1~/~94.3  & 42.8~/~91.9  & 28.3~/~94.5  & 33.8~/~92.2  & 24.7~/~94.6  & 29.0~/~93.9  & 35.4~/~92.5  & \underline{\textbf{31.4}}~/~93.4    \\ %\cline{2-10} 
                       & KL Matching           & 55.6~/~81.4  & 64.2~/~82.5  & 53.6~/~84.6  & 61.3~/~78.9  & 54.0~/~83.1  & 58.5~/~79.8  & 63.8~/~79.9  & 58.7~/~81.4    \\ %\cline{2-10} 
                       & Mahalanobis           & 41.1~/~91.2  & 59.4~/~87.2  & 43.6~/~91.1  & 56.8~/~86.1  & 48.8~/~89.9  & 56.8~/~87.3  & 58.2~/~86.5  & 52.1~/~88.5    \\ %\cline{2-10} 
                       & MaxLog                & 40.1~/~92.0  & 50.7~/~89.7  & 35.4~/~93.2  & 48.6~/~88.5  & 38.9~/~92.3  & 50.5~/~88.6  & 51.3~/~88.7  & 45.1~/~90.4    \\ %\cline{2-10} 
                       & Energy                & 37.1~/~92.4  & 49.0~/~89.8  & 32.5~/~93.6  & 46.3~/~88.7  & 36.4~/~92.5  & 48.7~/~88.7  & 49.6~/~88.9  & 42.8~/~90.7    \\ %\cline{2-10} 
                       & ViM                   & 28.5~/~94.1  & 48.2~/~90.0  & 25.1~/~95.3  & 36.8~/~91.7  & 24.5~/~95.2  & 53.4~/~88.9  & 45.7~/~90.0  & 37.4~/~92.2    \\ %\cline{2-10} 
                       & ReAct \(\tau\)=1      & 36.9~/~99.4  & 49.2~/~97.0  & 32.1~/~92.0  & 46.6~/~95.2  & 36.2~/~92.6  & 49.6~/~99.1  & 49.1~/~89.1  & 42.8~/~94.9    \\ %\cline{2-10} 
                       & DICE \(\rho\)=0.08       & 37.2~/~91.3  & 47.9~/~89.4  & 32.0~/~93.2  & 44.6~/~88.4  & 35.3~/~92.2  & 50.3~/~87.9  & 53.2~/~85.1  & 42.9~/~89.6    \\ \cmidrule{1-10}

\multirow{10}{*}{SWIN-T} 
                       & MSP                   & 64.7~/~85.8  & 69.4~/~84.7  & 66.9~/~85.4  & 68.2~/~84.0  & 68.3~/~85.0  & 72.4~/~83.2  & 74.4~/~80.1  & 69.2~/~84.0    \\ %\cline{2-10}
                       & ODIN                  & 43.1~/~91.6  & 56.4~/~88.4  & 46.7~/~90.6  & 51.6~/~88.0  & 50.9~/~89.1  & 61.2~/~86.0  & 63.1~/~83.3  & 53.3~/~88.2    \\ %\cline{2-10}
                       & OpenMax               & 51.7~/~90.2  & 55.4~/~90.0  & 53.1~/~91.1  & 50.9~/~90.1  & 42.9~/~91.9  & 49.9~/~90.9  & 53.7~/~87.8  & \underline{\textbf{51.1}}~/~90.3    \\ %\cline{2-10}
                       & KL Matching           & 62.7~/~84.2  & 67.6~/~83.2  & 62.5~/~84.1  & 65.6~/~81.9  & 63.9~/~82.9  & 69.6~/~80.3  & 71.8~/~78.0  & 66.3~/~82.1    \\ %\cline{2-10}
                       & Mahalanobis           & 63.0~/~86.2  & 65.8~/~85.8  & 55.8~/~89.1  & 67.6~/~84.4  & 54.8~/~88.2  & 70.6~/~83.1  & 67.1~/~82.2  & 63.5~/~85.6    \\ %\cline{2-10}
                       & MaxLog                & 45.5~/~91.9  & 58.3~/~88.4  & 49.4~/~90.5  & 54.2~/~87.7  & 54.2~/~88.8  & 64.2~/~85.5  & 65.5~/~83.1  & 55.9~/~88.0    \\ %\cline{2-10}
                       & Energy                & 43.6~/~92.2  & 58.0~/~88.4  & 48.0~/~90.7  & 53.1~/~87.8  & 53.3~/~88.9  & 63.8~/~85.5  & 65.2~/~83.1  & 55.0~/~88.1    \\ %\cline{2-10}
                       & ViM                   & 43.4~/~92.4  & 59.7~/~88.3  & 35.5~/~93.0  & 52.7~/~87.7  & 51.3~/~89.2  & 74.3~/~82.5  & 64.3~/~83.4  & 54.5~/~88.1    \\ %\cline{2-10}
                       & ReAct \(\tau\)=0.75   & 56.0~/~88.8  & 64.4~/~86.3  & 69.2~/~85.1  & 67.1~/~84.2  & 62.0~/~86.9  & 64.9~/~85.1  & 68.1~/~84.0  & 64.5~/~85.8    \\ %\cline{2-10}
                       & DICE \(\rho\)=0.1      & 58.6~/~88.8  & 69.6~/~84.6  & 79.2~/~81.1  & 74.0~/~81.5  & 66.9~/~85.2  & 67.0~/~84.0  & 72.9~/~82.1  & 69.7~/~83.9    \\ \bottomrule

\end{tabular}
}

\end{table}

\begin{table}[ht]
\setlength{\tabcolsep}{8pt}
\caption{Detailed performance for Non-Food dataset expressed by FPR95~/~AUROC. Best results are represented by \underline{\textbf{FPR95}}.}
\label{tab:OODNF}
\centering
\resizebox{1.0\textwidth}{!}{ % Resize table to fit text width
\begin{tabular}{c|c|c|c|c|c|c|c}
\toprule
\textbf{Model} & \textbf{Method} & \textbf{CIFAR10} & \textbf{SVHN} & \textbf{iSUN} & \textbf{places365} & \textbf{ImageNet1k} & \textbf{Average} \\ \midrule
\multirow{10}{*}{RN18} 
                       & MSP                   & 31.2~/~94.5  & 17.0~/~97.1  & 30.4~/~94.6  & 44.8~/~91.2  & 60.5~/~86.3  & 36.8~/~92.7  \\ %\cline{2-8}
                       & ODIN                  & 0.7~/~99.8   & 0~/~100     & 3.6~/~99.2  & 18.2~/~95.7  & 31.8~/~92.0  & 10.9~/~97.3  \\ %\cline{2-8}
                       & OpenMax               & 97.4~/~90.0  & 97.3~/~91.6 & 97.5~/~89.5 & 95.1~/~88.5  & 82.6~/~88.8  & 94.0~/~89.7  \\ %\cline{2-8}
                       & KL Matching           & 42.5~/~93.4  & 27.7~/~95.9 & 38.1~/~93.6 & 49.7~/~90.2  & 58.9~/~85.7  & 43.4~/~91.7  \\ %\cline{2-8}
                       & Mahalanobis           & 88.1~/~75.4  & 87.4~/~83.3 & 91.5~/~71.4 & 89.5~/~65.4  & 74.8~/~76.4  & 86.2~/~74.4  \\ %\cline{2-8}
                       & MaxLog                & 7.3~/~98.6   & 2.2~/~99.4  & 7.5~/~98.7  & 19.9~/~96.3  & 43.1~/~91.3  & 16.0~/~96.8  \\ %\cline{2-8}
                       & Energy                & 5.5~/~98.9   & 1.4~/~99.6  & 5.6~/~99.0  & 17.3~/~96.7  & 41.2~/~91.5  & 14.2~/~97.1  \\ %\cline{2-8}
                       & ViM                   & 0.3~/~99.7   & 0.4~/~99.7  & 0.3~/~99.6  & 1.8~/~99.3   & 3.6~/~98.8   & \underline{\textbf{1.3}}~/~99.4  \\ %\cline{2-8}
                       & ReAct \(\tau\)=1.75   & 3.5~/~99.2   & 0.2~/~99.9  & 3.4~/~99.3  & 10.5~/~97.9  & 27.1~/~94.8  & 8.9~/~98.2   \\ %\cline{2-8}
                       & DICE \(\rho\)=0.01        & 4.6~/~99.1   & 1.3~/~99.7  & 4.5~/~99.2  & 17.6~/~96.6  & 41.3~/~91.4  & 13.8~/~97.2  \\ \midrule

\multirow{10}{*}{RN50} 
                       & MSP                   & 38.6~/~93.0   & 20.3~/~96.7  & 35.9~/~93.8  & 45.0~/~91.6  & 56.1~/~88.7  & 39.2~/~92.8  \\ %\cline{2-8}
                       & ODIN                  & 6.9~/~98.6    & 0.0~/~99.9   & 14.0~/~97.4  & 39.0~/~91.4  & 40.8~/~90.6  & 20.1~/~95.6  \\ %\cline{2-8}
                       & OpenMax               & 63.7~/~90.8   & 59.1~/~92.3  & 54.4~/~92.8  & 56.0~/~92.1  & 56.4~/~91.4  & 57.9~/~91.9  \\ %\cline{2-8}
                       & KL Matching           & 39.8~/~92.8   & 21.1~/~96.7  & 36.0~/~93.9  & 43.8~/~91.9  & 54.2~/~88.8  & 39.0~/~92.8  \\ %\cline{2-8}
                       & Mahalanobis           & 42.6~/~92.9   & 28.1~/~95.4  & 38.2~/~94.0  & 27.2~/~95.3  & 36.6~/~92.7  & 34.5~/~94.1  \\ %\cline{2-8}
                       & MaxLog                & 19.6~/~96.7   & 5.9~/~98.8   & 17.2~/~97.1  & 25.7~/~95.3  & 39.7~/~92.5  & 21.6~/~96.1  \\ %\cline{2-8}
                       & Energy                & 18.3~/~96.9   & 5.6~/~98.8   & 15.8~/~97.2  & 24.5~/~95.5  & 38.6~/~92.6  & 20.6~/~96.2  \\ %\cline{2-8}
                       & ViM                   & 0.1~/~99.9    & 0.0~/~100.0  & 0.1~/~99.9   & 2.9~/~99.2   & 8.7~/~98.3   & \underline{\textbf{2.4}}~/~99.5  \\ %\cline{2-8}
                       & ReAct \(\tau\)=1       & 13.2~/~97.7   & 6.0~/~98.9   & 10.3~/~98.2  & 14.0~/~97.4  & 29.1~/~94.7  & 14.5~/~97.4  \\ %\cline{2-8}
                       & DICE \(\rho\)=0.05        & 16.5~/~97.1   & 2.4~/~99.5   & 13.7~/~97.5  & 28.2~/~94.7  & 39.1~/~92.3  & 20.0~/~96.2  \\ \midrule

\multirow{10}{*}{ViT-B} 
                       & MSP                   & 0.2~/~100.0    & 0.2~/~100.0  & 0.2~/~99.9  & 1.5~/~99.6   & 3.2~/~99.3   & 1.1~/~99.8    \\ %\cline{2-8}
                       & ODIN                  & 0.0~/~100.0   & 0.0~/~100.0   & 0.0~/~100.0  & 0.8~/~99.8   & 1.1~/~99.7   & 0.4~/~99.9    \\ %\cline{2-8}
                       & OpenMax               & 0.1~/~98.3   & 0.1~/~99.0  & 0.1~/~98.5  & 0.7~/~98.4   & 1.6~/~98.2   & 0.5~/~98.5    \\ %\cline{2-8}
                       & KL Matching           & 0.3~/~99.7   & 0.4~/~99.7  & 0.3~/~99.6  & 1.8~/~99.3   & 3.6~/~98.8   & 1.3~/~99.4    \\ %\cline{2-8}
                       & Mahalanobis           & 28.8~/~96.2  & 43.5~/~94.6 & 23.7~/~96.6 & 16.7~/~97.3  & 19.8~/~96.8  & 26.5~/~96.3   \\ %\cline{2-8}
                       & MaxLog                & 0.0~/~100.0   & 0.0~/~100.0   & 0.0~/~100.0  & 0.7~/~99.8   & 1.1~/~99.7   & 0.4~/~99.9    \\ %\cline{2-8}
                       & Energy                & 0.0~/~100.0   & 0.0~/~100.0   & 0.0~/~100.0  & 0.6~/~99.8   & 1.0~/~99.7   & 0.3~/~99.9    \\ %\cline{2-8}
                       & ViM                   & 0.0~/~100.0   & 0.0~/~100.0   & 0.0~/~100.0  & 0.5~/~99.9   & 0.7~/~99.8   & \underline{\textbf{0.2}}~/~99.9  \\ %\cline{2-8}
                       & ReAct \(\tau\)=1    & 0.0~/~100.0   & 0.0~/~100.0   & 0.0~/~100.0  & 0.6~/~97.0   & 1.0~/~99.3   & 0.3~/~99.3    \\ %\cline{2-8}
                       & DICE \(\rho\)=0.08       & 0.0~/~100.0   & 0.0~/~100.0   & 0.0~/~100.0  & 0.5~/~99.8   & 0.8~/~99.7   & 0.3~/~99.9    \\ \midrule

\multirow{10}{*}{SWIN-T} 
                       & MSP                   & 16.2~/~97.5   & 3.8~/~99.2   & 10.7~/~98.3  & 26.9~/~95.7   & 39.6~/~92.9   & 19.4~/~96.7    \\ %\cline{2-8}
                       & ODIN                  & 0.6~/~99.8    & 0.1~/~100.0  & 0.2~/~99.9   & 3.4~/~99.2    & 11.5~/~97.7   & 3.2~/~99.3    \\ %\cline{2-8}
                       & OpenMax               & 72.8~/~92.4   & 92.8~/~91.8  & 83.4~/~91.1  & 39.2~/~95.1   & 43.9~/~94.4   & 66.4~/~93.0    \\ %\cline{2-8}
                       & KL Matching           & 16.9~/~97.2   & 4.5~/~98.6   & 13.4~/~97.5  & 23.2~/~95.9   & 34.7~/~92.9   & 18.6~/~96.4    \\ %\cline{2-8}
                       & Mahalanobis           & 42.9~/~93.4   & 75.4~/~86.4  & 46.8~/~93.3  & 24.0~/~95.7   & 34.2~/~92.9   & 44.7~/~92.4    \\ %\cline{2-8}
                       & MaxLog                & 0.7~/~99.8    & 0.1~/~100.0  & 0.3~/~99.9   & 4.0~/~99.2    & 13.3~/~97.6   & 3.7~/~99.3    \\ %\cline{2-8}
                       & Energy                & 0.4~/~99.9    & 0.1~/~100.0  & 0.2~/~99.9   & 3.3~/~99.3    & 11.7~/~97.8   & 3.2~/~99.4    \\ %\cline{2-8}
                       & ViM                   & 0.0~/~100.0   & 0.0~/~100.0  & 0.0~/~100.0  & 0.8~/~99.8    & 2.7~/~99.4    & \underline{\textbf{0.7}}~/~99.8  \\ %\cline{2-8}
                       & ReAct \(\tau\)=0.75   & 5.4~/~98.7    & 0.9~/~99.7   & 3.4~/~99.0   & 12.8~/~97.6   & 28.8~/~94.8   & 10.2~/~97.9   \\ %\cline{2-8}
                       & DICE \(\rho\)=0.1      & 18.6~/~96.9   & 5.0~/~98.2   & 11.2~/~97.6  & 15.8~/~97.1   & 38.5~/~92.9   & 17.8~/~96.5   \\ \bottomrule

\end{tabular}
}
\end{table}
The average performance results across the food-related and non-food-related datasets are presented in \Cref{fig:all_models}.
We additionally provide the detailed results for each architecture and method for food OOD datasets in \Cref{tab:OODFOOD} and non-food OOD datasets in \Cref{tab:OODNF}. 
Due to the poorer performance of models trained from scratch, we instead report the detailed results of ResNet-18-S and ResNet-50-S in the Supplementary Materials: Experiments. 

Overall, we observed that all methods evaluated gave the best performance with the ViT-B model, which was also the model with the highest ID accuracy (\Cref{tab:Training specifics}), for both food and non-food OOD datasets, as seen in \Cref{fig:vitb}. 
Particularly, OpenMax \cite{bendale2016towards} and ViM \cite{wang2022vim} achieve the lowest FPR at TPR 95\% with values of 31.4, and 37.4 in food OOD datasets (\Cref{tab:OODFOOD}).
On the other hand, for non-food OOD datasets, most methods excelled with FPR95 between 0.2-1.3, with the exception of the Mahalanobis \cite{lee2018simple}, which had a relatively higher average FPR95 of 26.5 (\Cref{tab:OODNF}). 
The Swin-T model also showed high performance, with the lowest FPR95 of 51.1 in food OOD detection using OpenMax, and 0.7 in non-food OOD detection using ViM as seen in \Cref{fig:swint}, and \Cref{tab:OODFOOD,tab:OODNF}.
However, these patterns did not extend consistently to the CNN-based architectures. 
With ResNet-18 and ResNet-50, ViM outperformed other methods by delivering the lowest FPR95 scores for both food and non-food OOD datasets.
In contrast, OpenMax underperformed with these CNN models.

The pre-trained models consistently outperformed those trained from scratch, likely due to the lower ID accuracy of the latter. 
In our runs, we also noted that even models pre-trained on ImageNet-1K, such as ResNet-18 and ResNet-50 with ViM, and Swin-T with ViM, MaxLog, and Energy, produced relatively low FPR95 scores for the OOD ImageNet-1K (non-food items) dataset. 
This potentially suggests that the dataset on which the model was pre-trained does not heavily impact OOD performance, as models trained from scratch, such as ResNet-18-S and ResNet-50-S, exhibited significantly less favorable FPR95 and AUROC scores for the same datasets.

The overall metrics indicate that OOD detection methods performed better on non-food datasets than on food datasets (\Cref{fig:all_models}).
Simpler scoring functions, such as Energy, MaxLog, and ODIN, proved particularly effective on non-food datasets (\Cref{tab:OODNF}) where the feature space is well-separated. 
%Similarly, ODIN, also showed strong performance on non-food datasets, demonstrating its robustness in distinguishing between ID and OOD data when the feature space is well-separated.
The better performance on non-food datasets is likely due to the clearer feature distinction between ID and OOD data, clearly visible in \Cref{fig1}.
However, certain methods, such as OpenMax and Mahalanobis, performed unexpectedly poorly on non-food items, with FPR95 values exceeding 70 in some cases (\Cref{fig:resnet18,fig:resnet18s,fig:resnet50s,fig:swint} and \Cref{tab:OODNF}). 
While these methods may work well in some scenarios, they are not universally reliable across all model architectures and dataset types. 
On the other hand, ViM showed consistent performance across different OOD datasets and model architectures, proving its adaptability.
This suggests that ViM effectively uses both class-agnostic and class-specific information, even in highly variable food-related datasets.

Mahalanobis, in particular, struggled across the board, especially with food-related datasets, which are inherently more challenging due to their fine-grained nature. 
%Its reliance on Gaussian assumptions in the feature space limits its effectiveness when dealing with complex, real-world data
On the other hand, techniques that involve activation rectification and pruning, such as ReAct and DICE, generally outperformed simpler scoring functions like Energy, with ReAct consistently outperforming DICE in all runs.
This further emphasizes the importance of reducing redundant activations. 
Differently from the original work \cite{sun2022dice}, DICE performed better using a smaller values of \(\rho\) for activation pruning.
%, while our experiments showed that using a smaller \(\rho\) led to better results in our settings.

Additionally, transformer models clearly outperformed CNN-based ones, as reported by recent works \cite{fort2021exploring}. 
This could be possibly attributed not only to their higher ID accuracy but also to their ability to better handle complex variations in data, owing to their more abstract and high-dimensional feature representations.

\section{Conclusion}
In this work, we assessed various post-hoc OOD detection methods within the domain of fine-grained food recognition. 
Our results showed that ViM \cite{wang2022vim}, performed better generally - not only did it stand out in recognizing non-foods but it was always among the top-3 methods when it came to detecting food as OOD. 
Additionally, we note that model architecture and ID accuracy have a great impact on OOD detection performances. 
This work underscores the need for continued research into specialized OOD detection strategies tailored to fine-grained food recognition, particularly as these models find increasing application in real-world systems such as automatic dietary assessments.

\subsubsection{Acknowledgements} 
This work was partly supported by the European Commission and the Swiss Confederation - State Secretariat for Education, Research and Innovation (SERI) within the projects 101057730 Mobile Artificial Intelligence Solution for Diabetes Adaptive Care (MELISSA) and 101080117 BETTER4U.

%
% ---- Bibliography ----
%
% BibTeX users should specify bibliography style 'splncs04'.
% References will then be sorted and formatted in the correct style.
%
\bibliographystyle{splncs04}
\bibliography{references}

\begin{thebibliography}{10}
\providecommand{\url}[1]{\texttt{#1}}
\providecommand{\urlprefix}{URL }
\providecommand{\doi}[1]{https://doi.org/#1}

\bibitem{abdur2023comparative}
Abdur~Rahman, L., Papathanail, I., Brigato, L., Mougiakakou, S.: A comparative analysis of sensor-, geometry-, and neural-based methods for food volume estimation. In: Proceedings of the 8th International Workshop on Multimedia Assisted Dietary Management. pp. 21--29 (2023)

\bibitem{ABDURRAHMAN202473}
{Abdur Rahman}, L., Papathanail, I., Brigato, L., Spanakis, E.K., Mougiakakou, S.: Chapter 6 - food recognition and nutritional apps. In: Klonoff, D.C., Kerr, D., Espinoza, J.C. (eds.) Diabetes Digital Health, Telehealth, and Artificial Intelligence, pp. 73--83. Academic Press (2024)

\bibitem{llama3modelcard}
AI@Meta: Llama 3 model card  (2024), \url{https://github.com/meta-llama/llama3/blob/main/MODEL_CARD.md}

\bibitem{ataguba2024african}
Ataguba, G., Ezekiel, R., Daniel, J., Ogbuju, E., Orji, R.: African foods for deep learning-based food recognition systems dataset. Data in Brief  \textbf{53},  110092 (2024)

\bibitem{bendale2016towards}
Bendale, A., Boult, T.E.: Towards open set deep networks. In: Proceedings of the IEEE conference on computer vision and pattern recognition. pp. 1563--1572 (2016)

\bibitem{bossard2014food}
Bossard, L., Guillaumin, M., Van~Gool, L.: Food-101--mining discriminative components with random forests. In: Computer vision--ECCV 2014: 13th European conference, zurich, Switzerland, September 6-12, 2014, proceedings, part VI 13. pp. 446--461. Springer (2014)

\bibitem{dehais2016food}
Dehais, J., Anthimopoulos, M., Mougiakakou, S.: Food image segmentation for dietary assessment. In: Proceedings of the 2nd international workshop on multimedia assisted dietary management. pp. 23--28 (2016)

\bibitem{imagenet_cvpr09}
Deng, J., Dong, W., Socher, R., Li, L.J., Li, K., Fei-Fei, L.: {ImageNet: A Large-Scale Hierarchical Image Database}. In: CVPR09 (2009)

\bibitem{dosovitskiy2020image}
Dosovitskiy, A., Beyer, L., Kolesnikov, A., Weissenborn, D., Zhai, X., Unterthiner, T., Dehghani, M., Minderer, M., Heigold, G., Gelly, S., et~al.: An image is worth 16x16 words: Transformers for image recognition at scale. arXiv preprint arXiv:2010.11929  (2020)

\bibitem{fort2021exploring}
Fort, S., Ren, J., Lakshminarayanan, B.: Exploring the limits of out-of-distribution detection. Advances in Neural Information Processing Systems  \textbf{34},  7068--7081 (2021)

\bibitem{he2016deep}
He, K., Zhang, X., Ren, S., Sun, J.: Deep residual learning for image recognition. In: Proceedings of the IEEE conference on computer vision and pattern recognition. pp. 770--778 (2016)

\bibitem{hendrycks2019scaling}
Hendrycks, D., Basart, S., Mazeika, M., Zou, A., Kwon, J., Mostajabi, M., Steinhardt, J., Song, D.: Scaling out-of-distribution detection for real-world settings. arXiv preprint arXiv:1911.11132  (2019)

\bibitem{hendrycks2016baseline}
Hendrycks, D., Gimpel, K.: A baseline for detecting misclassified and out-of-distribution examples in neural networks. arXiv preprint arXiv:1610.02136  (2016)

\bibitem{hsu2020generalized}
Hsu, Y.C., Shen, Y., Jin, H., Kira, Z.: Generalized odin: Detecting out-of-distribution image without learning from out-of-distribution data. In: Proceedings of the IEEE/CVF conference on computer vision and pattern recognition. pp. 10951--10960 (2020)

\bibitem{kaur2019foodx}
Kaur, P., Sikka, K., Wang, W., Belongie, S., Divakaran, A.: Foodx-251: a dataset for fine-grained food classification. arXiv preprint arXiv:1907.06167  (2019)

\bibitem{krizhevsky2009learning}
Krizhevsky, A., Hinton, G., et~al.: Learning multiple layers of features from tiny images  (2009)

\bibitem{lee2017training}
Lee, K., Lee, H., Lee, K., Shin, J.: Training confidence-calibrated classifiers for detecting out-of-distribution samples. arXiv preprint arXiv:1711.09325  (2017)

\bibitem{lee2018simple}
Lee, K., Lee, K., Lee, H., Shin, J.: A simple unified framework for detecting out-of-distribution samples and adversarial attacks. Advances in neural information processing systems  \textbf{31} (2018)

\bibitem{liang2017enhancing}
Liang, S., Li, Y., Srikant, R.: Enhancing the reliability of out-of-distribution image detection in neural networks. arXiv preprint arXiv:1706.02690  (2017)

\bibitem{liu2020energy}
Liu, W., Wang, X., Owens, J., Li, Y.: Energy-based out-of-distribution detection. Advances in neural information processing systems  \textbf{33},  21464--21475 (2020)

\bibitem{liu2021Swin}
Liu, Z., Lin, Y., Cao, Y., Hu, H., Wei, Y., Zhang, Z., Lin, S., Guo, B.: Swin transformer: Hierarchical vision transformer using shifted windows. In: Proceedings of the IEEE/CVF International Conference on Computer Vision (ICCV) (2021)

\bibitem{lu2020gofoodtm}
Lu, Y., Stathopoulou, T., Vasiloglou, M.F., Pinault, L.F., Kiley, C., Spanakis, E.K., Mougiakakou, S.: gofoodtm: an artificial intelligence system for dietary assessment. Sensors  \textbf{20}(15), ~4283 (2020)

\bibitem{min2020isia}
Min, W., Liu, L., Wang, Z., Luo, Z., Wei, X., Wei, X., Jiang, S.: Isia food-500: A dataset for large-scale food recognition via stacked global-local attention network. In: Proceedings of the 28th ACM International Conference on Multimedia. pp. 393--401 (2020)

\bibitem{min2023large}
Min, W., Wang, Z., Liu, Y., Luo, M., Kang, L., Wei, X., Wei, X., Jiang, S.: Large scale visual food recognition. IEEE Transactions on Pattern Analysis and Machine Intelligence  \textbf{45}(8),  9932--9949 (2023)

\bibitem{netzer2011reading}
Netzer, Y., Wang, T., Coates, A., Bissacco, A., Wu, B., Ng, A.Y., et~al.: Reading digits in natural images with unsupervised feature learning. In: NIPS workshop on deep learning and unsupervised feature learning. vol.~2011, p.~4. Granada (2011)

\bibitem{oh2024we}
Oh, J.H., Falahkheirkhah, K., Bhargava, R.: Are we ready for out-of-distribution detection in digital pathology? In: International Conference on Medical Image Computing and Computer-Assisted Intervention. pp. 78--89. Springer (2024)

\bibitem{uecfoodpixcomplete}
Okamoto, K., Yanai, K.: {UEC-FoodPIX Complete}: A large-scale food image segmentation dataset. In: Proc. of ICPR Workshop on Multimedia Assisted Dietary Management(MADiMa) (2021)

\bibitem{panagiotou2023complete}
Panagiotou, M., Papathanail, I., Abdur~Rahman, L., Brigato, L., Bez, N.S., Vasiloglou, M.F., Stathopoulou, T., de~Galan, B.E., Pedersen-Bjergaard, U., van~der Horst, K., et~al.: A complete ai-based system for dietary assessment and personalized insulin adjustment in type 1 diabetes self-management. In: International Conference on Computer Analysis of Images and Patterns. pp. 77--86. Springer (2023)

\bibitem{pandey2022object}
Pandey, D., Parmar, P., Toshniwal, G., Goel, M., Agrawal, V., Dhiman, S., Gupta, L., Bagler, G.: Object detection in indian food platters using transfer learning with yolov4. In: 2022 IEEE 38th International conference on data engineering workshops (ICDEW). pp. 101--106. IEEE (2022)

\bibitem{papathanail2023nutritional}
Papathanail, I., Abdur~Rahman, L., Brigato, L., Bez, N.S., Vasiloglou, M.F., van~der Horst, K., Mougiakakou, S.: The nutritional content of meal images in free-living conditions—automatic assessment with gofoodtm. Nutrients  \textbf{15}(17), ~3835 (2023)

\bibitem{papathanail2021evaluation}
Papathanail, I., Br{\"u}hlmann, J., Vasiloglou, M.F., Stathopoulou, T., Exadaktylos, A.K., Stanga, Z., M{\"u}nzer, T., Mougiakakou, S.: Evaluation of a novel artificial intelligence system to monitor and assess energy and macronutrient intake in hospitalised older patients. Nutrients  \textbf{13}(12), ~4539 (2021)

\bibitem{papathanail2022feasibility}
Papathanail, I., Vasiloglou, M.F., Stathopoulou, T., Ghosh, A., Baumann, M., Faeh, D., Mougiakakou, S.: A feasibility study to assess mediterranean diet adherence using an ai-powered system. Scientific reports  \textbf{12}(1),  17008 (2022)

\bibitem{ravelli2020traditional}
Ravelli, M.N., Schoeller, D.A.: Traditional self-reported dietary instruments are prone to inaccuracies and new approaches are needed. Frontiers in nutrition  \textbf{7}, ~90 (2020)

\bibitem{sun2021react}
Sun, Y., Guo, C., Li, Y.: React: Out-of-distribution detection with rectified activations. Advances in Neural Information Processing Systems  \textbf{34},  144--157 (2021)

\bibitem{sun2022dice}
Sun, Y., Li, Y.: Dice: Leveraging sparsification for out-of-distribution detection. In: European Conference on Computer Vision. pp. 691--708. Springer (2022)

\bibitem{termritthikun2018nu}
Termritthikun, C., Kanprachar, S.: {Nu-ResNet: Deep residual networks for Thai food image recognition}. Journal of Telecommunication, Electronic and Computer Engineering (JTEC)  \textbf{10}(1-4),  29--33 (2018)

\bibitem{vasiloglou2018comparative}
Vasiloglou, M.F., Mougiakakou, S., Aubry, E., Bokelmann, A., Fricker, R., Gomes, F., Guntermann, C., Meyer, A., Studerus, D., Stanga, Z.: A comparative study on carbohydrate estimation: Gocarb vs. dietitians. Nutrients  \textbf{10}(6), ~741 (2018)

\bibitem{wang2022vim}
Wang, H., Li, Z., Feng, L., Zhang, W.: Vim: Out-of-distribution with virtual-logit matching. In: Proceedings of the IEEE/CVF conference on computer vision and pattern recognition. pp. 4921--4930 (2022)

\bibitem{who_emro}
{World Health Organization}: {Unhealthy diets}. \url{https://www.emro.who.int/noncommunicable-diseases/causes/unhealthy-diets.html}

\bibitem{who_ncd}
{World Health Organization}: {Noncommunicable diseases}. \url{https://www.who.int/news-room/fact-sheets/detail/noncommunicable-diseases} (September 2023)

\bibitem{xu2015turkergaze}
Xu, P., Ehinger, K.A., Zhang, Y., Finkelstein, A., Kulkarni, S.R., Xiao, J.: Turkergaze: Crowdsourcing saliency with webcam based eye tracking. arXiv preprint arXiv:1504.06755  (2015)

\bibitem{yang2024generalized}
Yang, J., Zhou, K., Li, Y., Liu, Z.: Generalized out-of-distribution detection: A survey. International Journal of Computer Vision pp. 1--28 (2024)

\bibitem{zhou2017places}
Zhou, B., Lapedriza, A., Khosla, A., Oliva, A., Torralba, A.: Places: A 10 million image database for scene recognition. IEEE Transactions on Pattern Analysis and Machine Intelligence  (2017)

\bibitem{zhou2021contrastive}
Zhou, W., Liu, F., Chen, M.: Contrastive out-of-distribution detection for pretrained transformers. arXiv preprint arXiv:2104.08812  (2021)

\end{thebibliography}

\end{document}